\documentclass[10pt,conference]{IEEEtran} 

\usepackage{algorithm,algorithmic,amsmath,amssymb,array,balance,booktabs,caption,cite,color,changes,enumerate,float,graphicx,multicol,multirow,siunitx,times}
\usepackage[bookmarks=true]{hyperref}
\definechangesauthor[color=purple, name={Laura Smith}]{LS}

\newcommand{\numstage}{S}
\newcommand{\horizon}{T}

\newcommand{\state}{\mathbf{s}}
\newcommand{\action}{\mathbf{a}}

\newcommand{\observation}{\mathbf{o}}

\newcommand{\traj}{\tau}
\newcommand{\KL}{\text{KL}}
\newcommand{\N}{\mathcal{N}}
\newcommand{\model}{\mathcal{M}}
\newcommand{\dataset}{\mathcal{D}}
\newcommand{\classifier}{\mathcal{C}}

\newcommand{\methodname}{automated visual instruction-following with demonstrations}
\newcommand{\metabbr}{AVID}

\title{
\metabbr: Learning Multi-Stage Tasks via\\Pixel-Level Translation of Human Videos%
}

\begin{document}

\author{\authorblockN{Laura Smith, Nikita Dhawan, Marvin Zhang, Pieter Abbeel, and Sergey Levine}
\authorblockA{Berkeley Artificial Intelligence Research, Berkeley, CA, 94720\\
Email: \texttt{smithlaura@berkeley.edu}}
}

\makeatletter
\let\@oldmaketitle\@maketitle%
\renewcommand{\@maketitle}{\@oldmaketitle%
    \centering
    \includegraphics[width=0.85\linewidth]{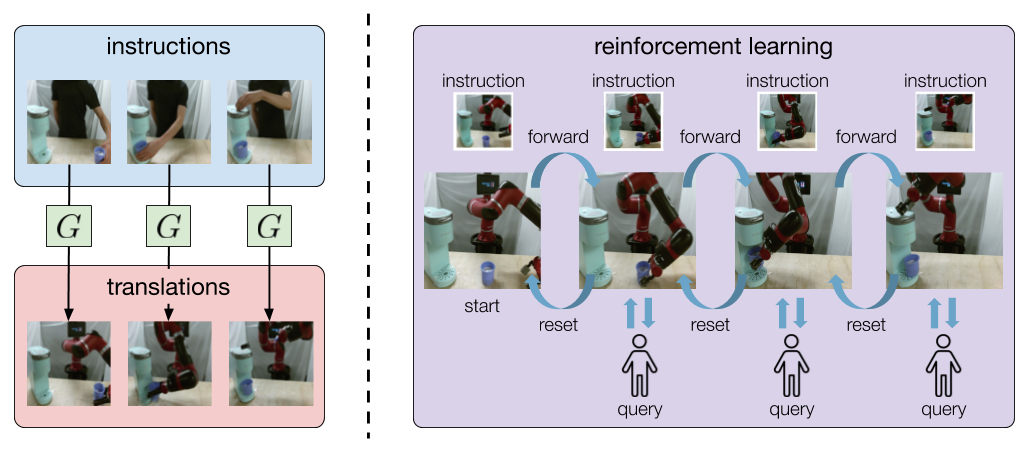}
    \captionof{figure}{Left: Human instructions for each stage (top) are translated at the pixel level into robot instructions (bottom) via CycleGAN. Right: The robot attempts the task stage-wise, automatically resetting and retrying until the instruction classifier signals success, prompting the human to confirm via key press. Algorithmic details are provided in \autoref{alg:overview}.}
    \label{fig:overview}
}
\makeatother

\maketitle

\thispagestyle{empty}
\pagestyle{empty}

\begin{abstract}
Robotic reinforcement learning (RL) holds the promise of enabling robots to learn complex behaviors through experience. However, realizing this promise for long-horizon tasks in the real world requires mechanisms to reduce human burden in terms of defining the task and scaffolding the learning process. In this paper, we study how these challenges can be alleviated with an automated robotic learning framework, in which multi-stage tasks are defined simply by providing videos of a human demonstrator and then learned autonomously by the robot from raw image observations. A central challenge in imitating human videos is the difference in appearance between the human and robot, which typically requires manual correspondence. We instead take an automated approach and perform pixel-level image translation via CycleGAN to convert the human demonstration into a video of a robot, which can then be used to construct a reward function for a model-based RL algorithm. The robot then learns the task one stage at a time, automatically learning how to reset each stage to retry it multiple times without human-provided resets. This makes the learning process largely automatic, from intuitive task specification via a video to automated training with minimal human intervention. We demonstrate that our approach is capable of learning complex tasks, such as operating a coffee machine, directly from raw image observations, requiring only 20 minutes to provide human demonstrations and about 180 minutes of robot interaction.
\end{abstract}

\section{Introduction}

Consider a robot learning to operate a coffee machine directly from visual inputs. This task is representative of many tasks we would like robots to learn in the real world in that, though the overall skill is complex, it can be broken down into multiple stages, each of which is simple to learn. In principle, reinforcement learning (RL) methods enable robots to learn skills such as this through their own experience. However, taking advantage of the natural stage-wise decomposition of the task is difficult with RL due to the challenges associated with defining a reward function for each stage, which often requires substantial human effort.

We may instead consider approaching this problem with imitation learning, as utilizing demonstrations is an effective way to convey intricate goals and to circumvent exploration challenges~\cite{robot-lfd}. One of the primary challenges with this approach is that robotic imitation learning typically relies on demonstrations provided on the robot itself. The means by which these demonstrations can be obtained are often laborious and limited in applicability~\cite{collab-manip,online-movement-adaptation,traj-keyframe,vr-teleop}. For instance, kinesthetic teaching requires a person to guide the robot by hand, which is not only unintuitive and physically demanding, but also introduces distractions or obstructions in image-based learning. Teleoperation, another common method for providing demonstrations, can require specialized hardware and generally demands considerable operator expertise.

Humans learn to imitate in a starkly different manner: through observation. In contrast to being remote controlled or physically maneuvered, humans can learn multi-stage tasks by simply \emph{watching} other people, imagining how they would perform the task themselves, and then practicing each stage of the task on their own. The question we ask in this work is: can we endow robots with the same ability?

We answer this question through the setting of learning from videos of human demonstrations. This setting eases the burden associated with providing demonstrations directly on the robot, as human demonstrations require comparatively minimal setup, hardware, and expertise. Thus, videos are often substantially easier for the user to provide, while opening up a much wider source of supervision for robots to learn skills. We still wish to leverage the demonstrations to provide natural, fine-grained guidance, such that the robot can learn effectively by exploiting the stage-wise nature of the task. However, as we learn from visual inputs, differences in appearance between the human and robot must be reconciled.

To tackle this challenge, we propose a robotic learning system that adopts an approach similar to the strategy described above: one of imagination and practice. Starting from a human demonstration, our method \emph{translates} this demonstration, at the pixel level, into images of the robot performing the task (see \autoref{fig:overview}). To remove the need for defining manual correspondences between the human and robot, we learn this translation via CycleGAN~\cite{cyclegan}, a framework for learning unpaired image-to-image translation. In order to handle multi-stage tasks, we reduce the human demonstration to a few \emph{instruction images}, denoting the stages of the task. These instruction images are translated using CycleGAN, which are then used to provide a reward for model-based RL, enabling the robot to practice the skill to learn its physical execution. This phase is largely automated, as the robot learns to reset each stage on its own to practice it multiple times. Human supervision is only needed in the form of selectively requested key presses, indicating success or failure, and a few manual resets. We demonstrate that this approach is capable of solving complex, long-horizon tasks with minimal human involvement, removing most of the human burden associated with instrumenting the task setup, manually resetting the environment, and supervising the learning process.

We name our method \methodname\ (\metabbr), and this method is the main contribution of our work. We evaluate \metabbr\ on two tasks using a Sawyer robot arm: operating a coffee machine and retrieving a cup from a drawer. \metabbr\ outperforms ablations and prior methods in terms of data efficiency and task success, learning coffee making with only 30 human demonstrations, amounting to 20 minutes of human demonstration time, and 180 minutes of robot interaction time with the environment.

\section{Related Work}
\label{sec:related}

\textbf{RL for Robotics.} RL is a powerful framework for enabling robots to autonomously learn a wide range of real-world skills (see, e.g.,~\cite{quadrupedal,policy-search,robot-rl,tactile-features,in-hand-manip,qtopt}). However, one of the key challenges in applying RL in the real world is in defining the reward function. This challenge is a major obstacle in deploying RL agents outside of controlled settings, as measuring rewards typically requires highly instrumented, and sometimes creative, setups such as motion capture and object tracking~\cite{em-rl,progressive-nets}, accelerometers to detect door opening~\cite{dagps}, or thermal cameras to detect pouring~\cite{pouring}. Our work, in line with robotic imitation learning approaches~\cite{robot-lfd}, circumvents the need for a reward function by using demonstrations to define the task.

For robotic applications, data efficiency is often another significant issue, and a number of approaches have been proposed to improve the data efficiency of RL algorithms. One popular approach is to use state representation learning to reduce the dimensionality of the data used for RL~\cite{srl-survey}. Another approach to data efficiency is to use model-based RL, which typically has been more efficient than model-free RL methods in practice~\cite{pilco,pets}. We frame representation learning and model learning in our work as a joint variational inference procedure, combining aspects of prior work to model real world images and sparse rewards~\cite{slac,vice}. However, we consider multi-stage tasks that are significantly more complex than those studied in these prior works.

\textbf{Learning multi-stage tasks.} A common paradigm for learning multi-stage tasks is to learn or plan separately for each stage, through methods such as chaining dynamic movement primitives~\cite{seq-mp} or linear-Gaussian controllers~\cite{reset-controller} or learning inverse dynamics models~\cite{rope-manip,htpi}. These works either assume access to reward functions for each stage or demonstrations consistent with the robot's observations and actions. Another complementary line of work studies automatically segmenting demonstrations into stages or subgoals using techniques such as movement primitive libraries~\cite{mp-demo} or generative models~\cite{causal-infogan}. We assume that the task stages are easy to manually specify by simply picking a few instruction images from the human demonstrations, thus, incorporating methods for automatic stage segmentation is left for future work. Instead, in this work we focus on reducing human burden during the learning process.

One way in which we reduce human burden is by enabling the robot to reset each stage of the task, rather than requiring constant manual resets. This is similar to~\cite{reset-controller}, which proposes an algorithm that uses learned resets and can solve a multi-stage toy wrench task. However, we evaluate on image-based tasks and we do not assume access to any reward functions. Thus, there is no mechanism for automatically testing whether a stage has been completed, and instead we use human key presses to indicate the success or failure of a stage. Using human feedback has been studied in several recent works~\cite{active,human-pref,vice-raq}, and we further decrease the data and human supervision requirements compared to these works, in order to learn multi-stage tasks in only a few hours of autonomous robot execution, a few manual resets, and less than 150 key presses.

\textbf{Learning from human demonstrations.} As mentioned earlier, prior methods for robotic imitation learning have typically used demonstrations consisting of observations and actions from the robot's observation and action space (see, e.g.,~\cite{learning-imitation,human-arm,comp-approaches,robot-lfd,mp-robot,obj-functions}). We instead study whether we can allow robots to learn from watching a human demonstrator, relaxing standard assumptions that the expert data are given using the robot’s own embodiment. Some works have studied this setting using explicit pose and object detection~\cite{kendama,learning-watching,activity-grammars,world-wide-web,semantic-rep}, essentially resolving the correspondence problem~\cite{correspondence} by instrumenting paired data collection or manually matching hand-specified key points. Other approaches have included predictive modeling~\cite{activity-forecasting,wwyd}, context translation~\cite{ifo,htpi}, learning reward representations~\cite{perceptual-rewards,tcn}, and meta-learning~\cite{daml}. In contrast to these works, we explicitly account for the change in embodiment through learned pixel-level translation, we evaluate on long-horizon multi-stage tasks, and we do not assume access to any demonstrations given directly on the robot. 

We evaluate the single-view version of time-contrastive networks (TCN)~\cite{tcn} on our tasks in \autoref{sec:experiments}. This prior method also handles embodiment changes, evaluates on visual robotic tasks, does not require demonstrations given on the robot, and has successfully learned pouring from videos of human pouring. TCN typically requires multiple views of human demonstrations, but we do not assume access to these in our problem setting, thus we compare to single-view TCN.

\section{Preliminaries}

To learn from human demonstrations, our method relies on several key steps: translating human videos, modeling robot images and actions, extracting instruction images, and stage-wise model-based RL. In this section, we review the first two steps for which we borrow techniques from prior work.

\subsection{Unsupervised Image-to-Image Translation}

We approach human to robot demonstration translation as an unsupervised image-to-image translation problem, where the goal is to map images from a source domain $X$ (human images, in our case) to a target domain $Y$ (robot images) in the absence of paired training data~\cite{cyclegan,unit}. Although the training data is not paired, we assume there exists a bijective mapping that can be uncovered in an unsupervised manner. This assumption is manifested in our experiments in \autoref{sec:experiments} as similarities in morphology and perspective between human and robot images. The approach we use to learn this translation is CycleGAN~\cite{cyclegan}, a framework which learns two mappings: $G:X\to Y$ translates source to target, and $F:Y\to X$ translates target to source. The translations are learned in order to fool discriminators $D_Y$ and $D_X$, which are trained to distinguish real images from translations in the target and source domains, respectively. This leads to a loss of the form
\[
\mathcal{L}_{\text{GAN}}(G,D_Y)=\mathbb{E}[\log D_Y(y)+\log(1-D_Y(G(x)))]\,,
\]
and similarly for $F$ and $D_X$, where $x$ and $y$ are source and target images drawn from the data distribution. An additional loss promotes cycle consistency between $G$ and $F$, i.e.,
\[
\mathcal{L}_{\text{cyc}}(G,F)=\mathbb{E}[\|x-F(G(x))\|_1+\|y-G(F(y))\|_1]\,.
\]
The overall training objective for CycleGAN is given by
\begin{align*}
\mathcal{L}_{\text{CG}}(G,F,D_X,D_Y)=\hspace{0.25em}&\mathcal{L}_{\text{GAN}}(G,D_Y)+\mathcal{L}_{\text{GAN}}(F,D_X)\\
                                                    &+\lambda\mathcal{L}_{\text{cyc}}(G,F)\,,
\end{align*}
where $\lambda$ is a hyperparameter.  Although this approach does not exploit the temporal information that is implicit in our demonstration videos, prior work has shown that CycleGAN can nevertheless successfully translate videos frame by frame, such as translating videos of horses into videos of zebras~\cite{cyclegan}.

\subsection{Structured Representation Learning}

\setcounter{figure}{1}
\begin{figure}[ht]%
    \centering
    \includegraphics[width=.9\linewidth]{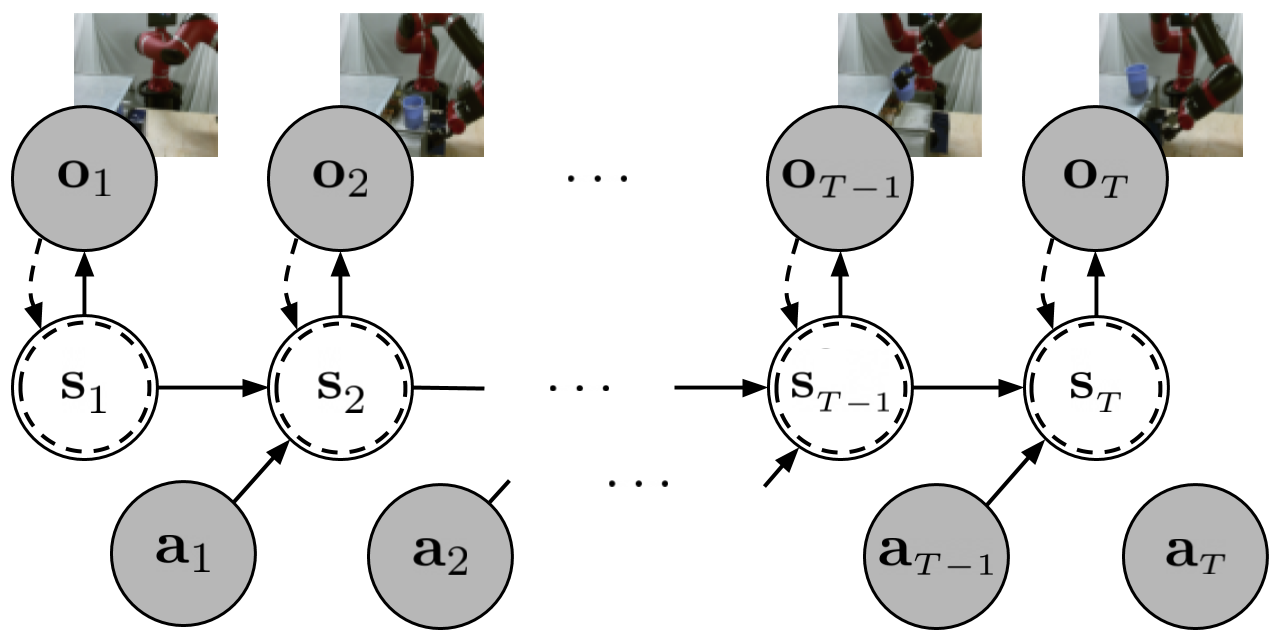}
    \caption{The latent variable model we use to represent robot images and actions, with the generative model depicted in solid lines and the variational family and encoder in dashed lines.}%
    \label{fig:model}
\end{figure}

Prior work has shown that, for control learning in image-based domains, state representation learning is an effective tool for improving data efficiency~\cite{srl-survey,solar,planet,slac}. We also use representation learning in our work, and similar to~\cite{slac}, we learn a latent state representation of our image observations by defining a probabilistic temporally-structured latent variable model. In particular, we assume the underlying state of the system $\state_t$ is unobserved but evolves as a function of the previous state and action, and we treat the robot images as observations $\observation_t$ of this state.
The generative model for $\state_t$ can be summarized as an initial state distribution and a learned neural network dynamics model, i.e.,
\begin{align*}
                           p(\state_1)&=\N\left(\state_1;0,\mathbf{I}\right)\,,\\
    p(\state_{t+1}|\state_t,\action_t)&=\N\left(\state_{t+1};\mu(\state_t,\action_t),\Sigma(\state_t,\action_t)\right)\,,
\end{align*}
where $\mu$ and $\Sigma$ are parameterized by neural networks. To complete the generative model, we learn a decoder $p(\observation_t|\state_t)$ which we represent as a convolutional neural network. In order to learn this model, we introduce a variational distribution $q(\state_{1:\horizon};\observation_{1:\horizon})$ which approximates the posterior $p(\state_{1:\horizon}|\observation_{1:\horizon},\action_{1:\horizon})$, where the $1:\horizon$ notation denotes entire trajectories. We use a mean field variational approximation
\[
q(\state_{1:\horizon};\observation_{1:\horizon})=\prod_tq(\state_t;\observation_t)\,,
\]
where the encoder $q(\state_t;\observation_t)$ is also a convolutional network. We jointly learn the parameters of $p$ and $q$ via maximization of the variational lower bound (ELBO), given by
\begin{align*}
    \text{ELBO}&[p,q]=\mathbb{E}_q[p(\observation_t|\state_t)]-D_\KL(q_{\state_1}(\cdot;\observation_1)\|p_{\state_1})\\
               &-\mathbb{E}_q\left[\sum_{t=1}^{\horizon-1}D_\KL(q_{\state_{t+1}}(\cdot;\observation_{t+1})\|p_{\state_{t+1}}(\cdot|\state_t,\action_t))\right]\,.
\end{align*}
The entire structured latent variable model is visualized in \autoref{fig:model}. As we explain in the next section, this model provides us with an encoder and dynamics model that we use for encoding instruction images and for stage-wise model-based planning in the learned latent space.

\section{Automated Visual Instruction-Following with Demonstrations}

In our problem setting, the task we want the robot to learn is specified by a set of human demonstration videos, each of which is a trajectory of image observations depicting the human performing the task. We do not assume access to demonstrations given through teleoperation or kinesthetic teaching, which often require specific hardware and human expertise. We also do not assume access to rewards provided through motion capture or other instrumented setups. Thus, our goal is to enable robotic learning while requiring minimal human effort, expertise, and instrumentation. Task specification via a human demonstration is a step toward this goal, and this section details other design choices in CycleGAN training and robot learning that reflect this goal.

\subsection{Translation for Goal Concept Acquisition}

\begin{figure}[ht]%
    \centering
    \includegraphics[width=.85\linewidth]{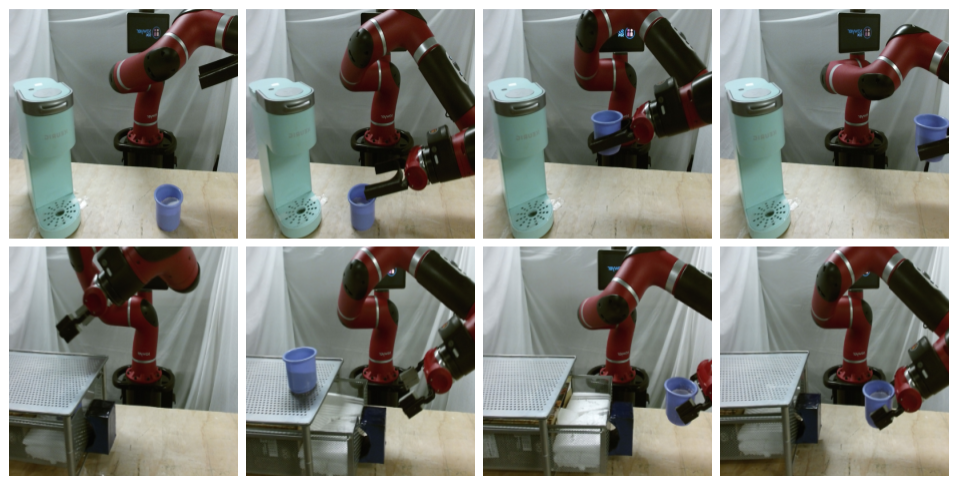}
    \caption{Examples of robot data collected to train the CycleGAN for coffee making (top) and cup retrieval (bottom). Though the robot moves randomly, we cover different settings by, e.g., giving the cup to the robot or opening the drawer.}%
    \label{fig:cyclegan}
\end{figure}

CycleGAN requires training data from both the source and target domains. We found that using diverse data was important for capturing a wide range of possible scenes and generating realistic translations. To that end, the training images we use from the human domain consist of both human demonstrations as well as a small amount of ``random'' data, in which the human moves around in the scene but does not specifically attempt the task. The training images from the robot domain consist entirely of the robot executing randomly sampled actions in a few different settings, and as shown in \autoref{fig:cyclegan}, in between settings we manually change the environment by shuffling objects or providing the robot with an item to hold. These types of human interventions do not require any specialized setups or technical ability from the human, and we found that a handful of interventions was sufficient for CycleGAN data collection (see \autoref{sec:experiments}). Thus, this overall procedure requires far less human expertise and effort compared to providing a set of robot demonstrations.

Once we have trained the CycleGAN model, the mapping $G$ provides a mechanism for automatically translating human demonstration videos to robot demonstration videos. Though we could use an imitation learning method that directly learns from the entire translated video, we show in \autoref{sec:experiments} that this approach performs poorly. Prior work in imitation learning has shown the efficacy of using select frames or states to imitate rather than using the entire demonstration~\cite{zsvi,htpi,selective-il}, and we employ the same approach by extracting and learning from key frames that correspond to the completion of stages. The resulting decoupling of goals from how they can be achieved is especially natural for multi-stage tasks, where achieving particular states that are necessary for overall success is much more important than precisely matching a particular trajectory.

Specifically, the \emph{instruction images} for stage $i$ are taken from the $t_i$-th frame of the translated videos, where the time steps $\{t_1,\ldots,t_\numstage\}$ are manually specified by the user. This process is easy for the human as we use a modest number of videos and, as mentioned, the time steps intuitively correspond to the completion of natural stages in the task (see \autoref{fig:overview}). As discussed in \autoref{sec:related}, future work could explore automatic stage discovery for more complex tasks, potentially via video segmentation methods~\cite{tap,tsc,perceptual-rewards}.

To specify a reward function for each stage, we train success classifiers for each stage $\{\classifier_1,\ldots,\classifier_\numstage\}$, where $\classifier_i$ is provided the instruction images at time step $t_i$ as positive examples and all other robot images as negative examples. In order to achieve greater data efficiency, we first encode the images using the encoder $q(\state_t;\observation_t)$ and use the learned latent state as input to the classifier. The log probabilities from the classifiers can then be used as a reward signal, similar to~\cite{vice,vice-raq}. Providing rewards in this way does not rely on any manually designed reward shaping or instrumentation. Furthermore, the use of learned classifiers allows for online refinement, an avenue for the human to provide feedback. As we discuss next, this setup lends itself naturally to an automated stage-wise model-based planning procedure.

\subsection{Model-Based RL with Instruction Images}

Though we could use any RL algorithm with our classifier-based reward, model-based RL has typically achieved greater data efficiency than model-free methods~\cite{pilco,pets}, and the data efficiency afforded by the combination of model-based RL and representation learning further reduces the human supervision required during learning. Thus, the RL procedure that we use is the latent space model-predictive control (MPC) method described in~\cite{solar}, which involves iteratively searching for an optimal sequence of actions using the cross-entropy method (CEM)~\cite{cem}, a sampling-based optimization procedure. The search is refined by evaluating model-generated trajectories in the latent space under the reward function, which is given by a classifier in our case. The robot then executes the first action of the optimized sequence. We encapsulate this procedure in the \texttt{MPC-CEM} subroutine in \autoref{alg:overview}.

When the robot is attempting stage $s$, the planner uses the log probability of $\classifier_s$ as the reward function and aims to surpass a classifier threshold $\alpha\in[0,1]$, which is a hyperparameter. If this threshold is not met, the planner automatically switches to $\classifier_{s-1}$, i.e., it attempts to reset to the beginning of the stage. This forward-reset behavior allows the robot to robustify its performance with very few human-provided resets, as the human only intervenes to fix problems, such as the cup falling over, rather than manually resetting every episode. The robot runs this forward-reset loop for a maximum of $K$ iterations, where $K$ is also a hyperparameter.

Should the threshold be met during planning, the robot will query the human user, at which point the user signals either success or failure through a key press. On failure, the robot switches to the reset behavior, and the loop continues. On success, the robot moves on to the next stage and repeats the same process, with $\classifier_{s+1}$ specifying the goal and $\classifier_s$ specifying the reset. This stage-wise learning avoids the compounding errors of trying to learn the entire task all at once. The full procedure is summarized in \autoref{alg:overview}.

\begin{algorithm}[H]
\caption{\metabbr\ RL Forward Pass}
\label{alg:overview}
\begin{algorithmic}[1]
\footnotesize
\REQUIRE  Pre-trained model $\model$ and $\{\classifier_i\}_{i=0}^{\numstage}$, initial robot data $\dataset$\\
\REQUIRE  Max attempts per stage $K$, classifier threshold $\alpha$
\STATE \texttt{stage} $\leftarrow1$
\WHILE{\texttt{stage} $\leq\numstage$} 
    \STATE \texttt{attempts} $\leftarrow0$
    \STATE \texttt{stage-completed} $\leftarrow$ \textit{false}
    \WHILE{\texttt{attempts} $<K$ \textbf{and not} \texttt{stage-completed}}
        \STATE $\traj\leftarrow$ \texttt{MPC-CEM}$(\model,\classifier_\texttt{stage})$; $\dataset\leftarrow\dataset\cup\traj$
        \STATE Increment \texttt{attempts}
        \IF{$\classifier_{\texttt{stage}}(\text{last observation of }\tau)> \alpha$}   
            \IF{human signals success}
                \STATE Add last observation of $\tau$ to goal images for \texttt{stage}
                \STATE \texttt{stage-completed} $\leftarrow$ \textit{true} \\
            \ELSE
                \STATE $\dataset\leftarrow\dataset~\cup$ \texttt{EXPLORE}$()$
                \STATE Train $\model$ and $\{\classifier_{i}\}_{i=0}^{\numstage}$ on $\dataset$ and goal images
            \ENDIF
        \STATE \texttt{attempts} $\leftarrow0$
        \ELSE 
            \STATE $\traj\leftarrow$ \texttt{MPC-CEM}$(\model,\classifier_{\texttt{stage}-1})$; $\dataset\leftarrow\dataset\cup\traj$
        \ENDIF
    \ENDWHILE
    \IF{\texttt{stage-completed}}
        \STATE Increment \texttt{stage}
    \ELSE
        \STATE Train $\model$ and $\{\classifier_{i}\}_{i=0}^{\numstage}$ on $\dataset$ and goal images\\
        \STATE \texttt{attempts} $\leftarrow0$
    \ENDIF
\ENDWHILE
\end{algorithmic}
\end{algorithm}

We include several important improvements on top of this procedure in order to achieve good performance. In line 13 of \autoref{alg:overview}, if the human signals failure, the \texttt{EXPLORE} subroutine is invoked, in which the robot moves randomly from the state it ended in and queries the human several times. This provides additional data within the critical area of what $\classifier_s$ believes is a success. In addition, in line 14, the current image is labeled as a negative for $\classifier_s$ and added to the dataset along with the images from random exploration. Similar to~\cite{vice,vice-raq}, further training $\classifier_s$ on this data improves its accuracy by combating the false positive problem and offsetting the artifacts in the translated images that $\classifier_s$ was initially trained with. Finally, to further automate the entire process and avoid manual human resets, after the robot completes the last stage, we run \autoref{alg:overview} in reverse in order to get back to the initial state. We repeat this entire process until the robot can reliably complete the entire task.

\section{Experiments}
\label{sec:experiments}

\begin{table*}[ht]
\centering
\resizebox{\textwidth}{!}{
\scriptsize
\begin{tabular}{lcccccccccc}
\toprule
                     &             & \multicolumn{3}{c}{\textbf{Demo supervision}}         & \multicolumn{6}{c}{\textbf{Data collected (\# images)}} \\
                                     \cmidrule(lr){3-5}                                      \cmidrule(lr){6-11}
                     &             & \textbf{Human}   & \multicolumn{2}{c}{\textbf{Robot}} & \multicolumn{3}{c}{\textbf{Coffee making}} & \multicolumn{3}{c}{\textbf{Cup retrieval}} \\
                                     \cmidrule(lr){3-3} \cmidrule(lr){4-5}                   \cmidrule(lr){6-8}                           \cmidrule(lr){9-11}
\textbf{Method}      & Stage-aware & Obs              & Obs        & Actions               & Demo & Pretraining & Online                & Demo & Pretraining & Online \\
\midrule
\metabbr\ (ours)     & \checkmark  & \checkmark       &            &                       & 900  & 8600        & 3015                  & 600  & 7500        & 1740 \\
Full-video ablation  &             & \checkmark       &            &                       & 900  & 15800       & 1350                  & 600  & 14700       & 1400 \\
Pixel-space ablation & \checkmark  & \checkmark       &            &                       & 900  & 15800       & 0                     & 600  & 14700       & 0 \\
TCN~\cite{tcn}       &             & \checkmark       &            &                       & 1395 & 14400       & 3060                  & 900  & 14400       & 1800 \\
BCO~\cite{bco}       &             &                  & \checkmark &                       & 900  & 15300       & 900                   & 900  & 14400       & 300 \\
Behavioral Cloning   &             &                  & \checkmark & \checkmark            & 900  & 0           & 0                     & 900  & 0           & 0 \\
\bottomrule
\end{tabular}%
}
\caption{Overview of assumptions and data requirements of the evaluated methods. ``Stage-aware" indicates whether the approach allows for leveraging the multi-stage nature of tasks, which only our method and the pixel-space ablation do. ``Demo supervision" describes the type of demonstration data used: all of our comparisons can learn from human demonstrations except BCO and behavioral cloning, which rely on teleoperated demonstrations. We also report the number of images used by each method broken down by demonstrations, pretraining data (including non-demonstration human data and random robot data), and online training data. Because pixel-space models are generally less data efficient than latent variable models, we collect additional pretraining data for all methods except for ours. Note that, analogous to prior work~\cite{vf}, the pixel-space ablation is not trained online because it is too computationally expensive, and behavioral cloning does not have an online training phase.
}
\label{tab:cmprs}%
\end{table*}%

We aim to answer the following through our experiments:
    \begin{enumerate}[(1)]
    \item Is \metabbr\ capable of solving temporally extended vision-based tasks directly from human demonstrations?
    \item What benefits, if any, do we gain from using instruction images and latent space planning?
    \item What costs, if any, do we incur from not having access to demonstrations given directly on the robot?
\end{enumerate}

\begin{figure*}
  \includegraphics[width=\textwidth]{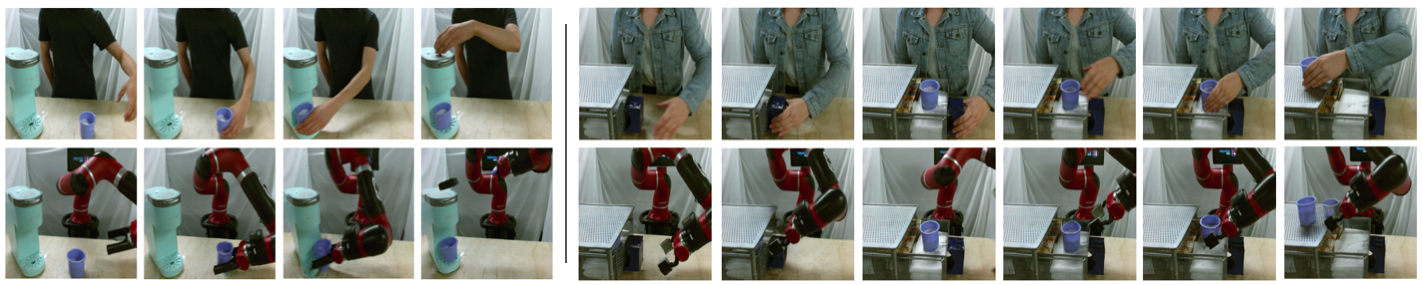}
  \caption{Sample sequence of instructions for each of the two tasks, coffee making (left) and cup retrieval (right). The instruction images are segmented from a human demonstration (top), then translated into the robot's domain (bottom) via CycleGAN. The stages for coffee making from left to right are: initial state, pick up the cup, place the cup in the coffee machine, and press the button on top of the machine. For the cup retrieval, the stages are: initial state, grasp the handle, open the drawer, lift the arm, pick up the cup, and place the cup on top. Note the artifacts in the generated translations, e.g., the displaced robot gripper in the last instruction image on the left and the deformed cup in the last image on the right.}
  \label{fig:allims}
\end{figure*}

To answer (1), we evaluate on two complex and temporally extended tasks (see \autoref{fig:allims}). For (2), we compare \metabbr\ to two ablations: one which learns from the entire translated demonstration and one which plans directly in pixel space. Finally, to answer (3), we evaluate how imitation learning fares on the same tasks when given access to ``oracle'' information, i.e., teleoperated robot demonstrations. We also analyze the human supervision burden of \metabbr\ (see \autoref{fig:humanfeedback}). A supplementary video depicting the learning process of \metabbr, as well as final performances for \metabbr\ and all of the comparisons, is available from the project website.\footnote{\tt \url{https://sites.google.com/view/rss20avid}}

\subsection{Comparisons}

We compare to the following methods:

\textbf{Behavioral cloning.} Behavioral cloning directly applies supervised learning on observation-action pairs from an expert to approximate the optimal policy. Although this ``oracle'' approach uses teleoperated demonstrations given on the robot, it is not designed to incorporate the notion of stages and is known to suffer from distributional shift arising from compounding errors~\cite{dagger}. Thus, this comparison serves to help determine what benefits \metabbr\ derives from stage-wise RL.

\textbf{Behavioral Cloning from [Robot] Observation (BCO)~\cite{bco}.} BCO also uses teleoperated demonstrations but only assumes access to observations. This method infers the actions from the observations using a learned inverse model, trained separately, and then performs behavioral cloning on the inferred actions. Note that our stage-wise training procedure, again, is not applicable. 

\textbf{Full-video ablation.} We consider an ablation of \metabbr\ that trains on entire demonstrations rather than just the instruction images. A natural way to learn from full videos is to use an imitation learning method that does not require expert actions~\cite{bco,fail,ilpo,daml,ifo,zsvi}. Here we again employ BCO, using translated human demonstrations rather than teleoperated demonstrations.

\textbf{Pixel-space ablation.} Deep visual foresight (DVF)~\cite{vf} is a visual model-based RL method that performs planning in pixel space instead of a learned latent space. We consider an ablation of \metabbr\ in which we use translated instruction images and our stage-wise training procedure but replace latent space planning with DVF. Note that this ablation is similar to \metabbr\ in that it uses stage-wise learning, learned resets, and human feedback.

\textbf{Time-contrastive networks~\cite{tcn}.} We evaluate the single-view version of TCN, which learns an embedding via a temporal consistency loss in order to generalize from human demonstrations to robot execution. TCN originally used PILQR for their RL algorithm~\cite{pilqr}, though as the authors highlight, any RL algorithm may be used with the learned embedding. For fairness and ease of comparison, we use the same RL subroutine as \metabbr. While \metabbr\ learns classifiers for the reward function, this comparison uses the negative Euclidean distance between TCN embeddings of the human demonstration and model-generated trajectories, decoded using $p(\observation_t|\state_t)$, as the reward function. We compare to the single-view version of TCN, as the multi-view version requires that the human demonstrations are filmed from multiple viewpoints, and we do not make the same assumption in our setup.

\subsection{Experimental Setup}

\begin{table*}[ht]
\centering
\resizebox{\textwidth}{!}{
\begin{tabular}{llcccccccc}
\toprule
                                          &                      & \multicolumn{3}{c}{\textbf{Coffee making}} & \multicolumn{5}{c}{\textbf{Cup retrieval}} \\
                                                                   \cmidrule(lr){3-5}                           \cmidrule(lr){6-10}
\textbf{Supervision}                      & \textbf{Method}      & Stage 1     & Stage 2   & Stage 3          & Stage 1     & Stage 2     & Stage 3     & Stage 4    & Stage 5 \\
\midrule
\multirow{4}{*}{Human Demos}              & \metabbr\ (ours)     & {\bf 100\%} & {\bf 80\%} & {\bf 80\%}      & {\bf 100\%} & {\bf 100\%} & {\bf 100\%} & {\bf 80\%} & {\bf 70\%} \\
                                          & Full-video ablation  & 70\%        & 10\%       & 0\%             & 0\%         & 0\%         & 0\%         & 0\%        & 0\% \\
                                          & Pixel-space ablation & 60\%        & 20\%       & 0\%             & 50\%        & 50\%        & 30\%        & 10\%       & 0\% \\
                                          & TCN~\cite{tcn}       & 10\%        & 10\%       & 0\%             & 60\%        & 20\%        & 0\%         & 0\%        & 0\% \\
\midrule
\multirow{2}{*}{Teleoperated Robot Demos} & BCO~\cite{bco}       & 80\%        & 30\%       & 0\%             & 30\%        & 10\%        & 0\%         & 0\%        & 0\% \\ 
                                          & Behavioral Cloning   & 90\%        & 90\%       & 90\%            & 100\%       & 100\%       & 60\%        & 60\%       & 40\% \\
\bottomrule
\end{tabular}%
}
\caption{We report success rates up to and including each stage for both tasks, over 10 trials. The top rows are methods that learn from human demonstrations, and we bold the best performance in this category. The bottom two rows are methods trained with direct access to demonstrations given on the real robot itself. \metabbr\ outperforms all other methods from human demonstrations, succeeding 8 times out of 10 on coffee making and 7 times out of 10 on cup retrieval, and even outperforms behavioral cloning from teleoperated demonstrations on the later stages of cup retrieval.}
\label{tab:results}%
\end{table*}%

We evaluate our method on two temporally-extended tasks from vision: operating a personal coffee machine and retrieving a cup from a closed drawer. These tasks illustrate that our method can learn to sequentially compose several skills by following a set of instructions extracted from a human demonstration. For all our experiments, we use end-effector velocity control on a 7 DoF Sawyer robotic manipulator, and our observations consist only of \mbox{64-by-64-by-3} RGB images.

For data collection, we film videos of human actions and random robot trajectories. This gives us data from the two domains of interest, and we train a CycleGAN using this data for \metabbr, the full-video ablation, and pixel-space ablation. The human demonstration videos are then reused to generate translated robot demonstrations. We also reuse the random robot trajectories to train the models needed for each method: the latent variable model for \metabbr, the dynamics model for the pixel-space ablation, the inverse models for the full-video ablation and BCO, and the embedding function for TCN. A breakdown of the amount and types of data used by each method for each of the tasks can be found in \autoref{tab:cmprs}.

\textbf{Coffee making.} The coffee making task has three stages as depicted in \autoref{fig:allims}. Starting with the cup on the table, the instructions are to pick up the cup, place the cup in the machine, and press the button on the top of the machine. For this task, we used 30 human demonstrations (900 images) along with 500 images of random human data, which took about 20 minutes. We also collected trajectories of the robot executing randomly sampled actions, with the human changing the setting 6 times by moving the cup and placing the cup in the robot's gripper. We set the classifier threshold $\alpha$ to 0.8 and the maximum number of attempts per stage $K$ to 3.

\textbf{Cup retrieval.} Cup retrieval has five stages, as shown in \autoref{fig:allims}: grasping the drawer handle, opening the drawer, moving the arm up and out of the way, picking up the cup, and placing the cup on top of the drawer. We found that the intermediate stage of moving the arm was important for good performance, as the model-based planning would otherwise bump into the drawer door. Providing this instruction was trivial with our setup, as we simply specified an additional instruction time step between the ``natural'' instructions. For this task, we used 20 human demonstrations (600 images) and 300 images of random human data, again amounting to about 20 minutes of human time. In collecting the random robot data, the human changed the setting 11 times by moving the cup, placing the cup in the robot's gripper, and opening the drawer. $\alpha$ and $K$ were again set to 0.8 and 3.

\subsection{Experimental Results}

We report success rates for both coffee making and cup retrieval in \autoref{tab:results}. Success rates are evaluated using 10 trials per method per task, with no human feedback or manual intervention. Success metrics are defined consistently across all methods and are strict, e.g., placing the cup on top of the drawer but toppling the cup over is a failure. The supplementary video shows representative evaluation trials, labeled with success and failure, for each method.\footnote{\tt \url{https://sites.google.com/view/rss20avid}} For the final stage of coffee making, as it is difficult to visually discern whether the robot has noticeably depressed the coffee machine button, success is instead defined as being within \SI{5}{\cm} above the button, such that a pre-programmed downward motion always successfully presses the button.

For both tasks, \metabbr\ achieves the best performance among the methods that use human demonstrations, consistently learning the earlier stages perfectly and suffering less from accumulating errors in the later stages. As seen in the supplementary video, the failures of \metabbr\ for both tasks correspond to generally good behavior with small, but significant, errors, such as knocking over or narrowly missing the cup. \metabbr\ makes constant use of automated resetting and retrying during both RL and the final evaluation, allowing for more robust behavior compared to the methods that cannot exploit the stage-wise decomposition of the tasks. Though the pixel-space ablation also incorporates resetting, this method is less successful than \metabbr. This may be due to the fact that the pixel-space model is not trained online or used for replanning as this is prohibitively expensive. This highlights the benefits of our latent variable model and planning procedure.

\begin{figure}[ht]%
    \raggedleft
    \includegraphics[width=.85\linewidth]{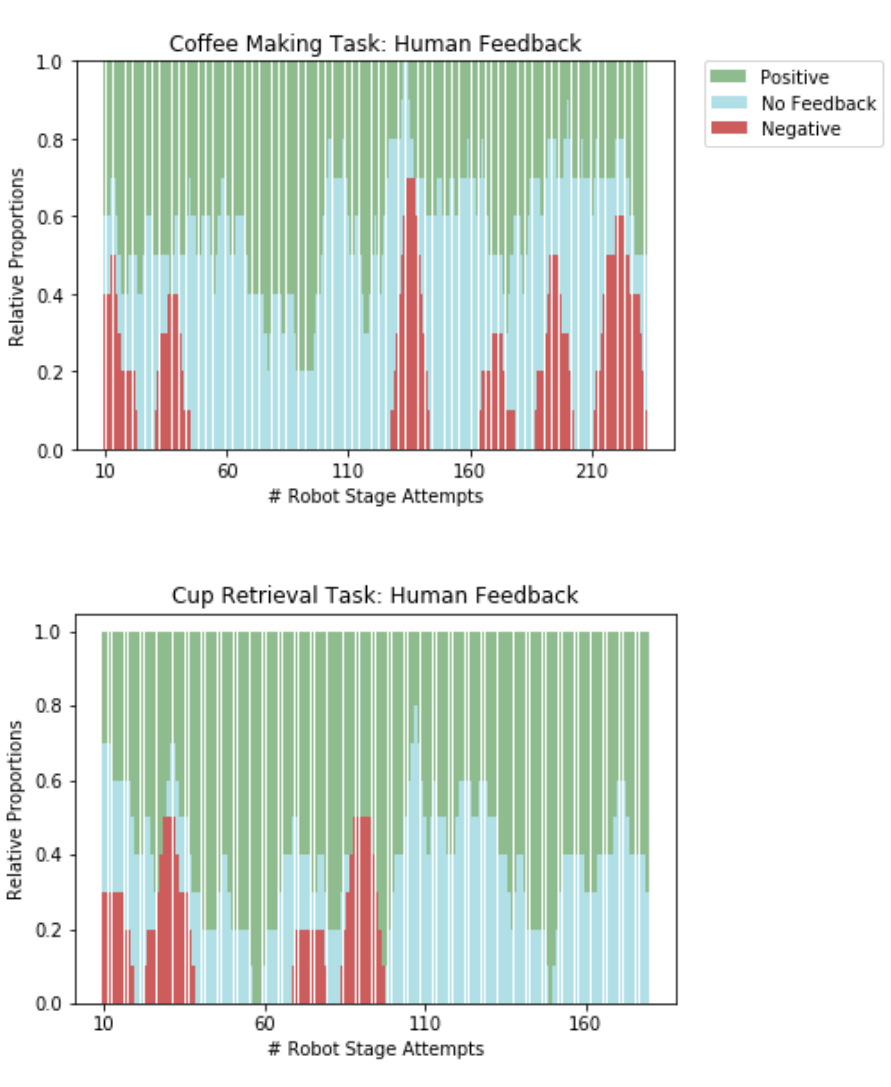}
    \caption{Visualizing human feedback during the learning process for coffee making (top) and cup retrieval (bottom). The x-axis is the total number of robot stage attempts, and the y-axis indicates the proportions of feedback types, smoothed over the ten most recent attempts. ``No feedback'' means that the classifier did not signal success and the robot automatically switched to resetting. Coffee making and cup retrieval used a total of 131 and 126 human key presses, respectively.}%
    \label{fig:humanfeedback}
\end{figure}

The full-video ablation can learn to pick up the cup 70\% of the time for the coffee making task, but it fails on the later stages and also fails entirely for cup retrieval. This indicates the difficulty of learning from full translated demonstrations due to the artifacts in the translations and the absence of stage-wise training. The supplementary video shows that, although the behavior of this method visually looks almost correct in many instances, it generally still fails in subtle ways. Finally, despite using publicly available code from the authors, we were unable to achieve good performance with TCN on coffee making, though TCN was moderately successful on the early stages of cup retrieval. We note that we used the single-view version of this method, and the authors also found that this version did not perform well for their tasks~\cite{tcn}. As previously mentioned, the multi-view version of TCN is not applicable in our case as we do not assume multiple views of human demonstrations at training time.

In order to understand how using translated human demonstrations compares to using real robot demonstrations, we evaluate BCO and behavioral cloning from demonstrations obtained through teleoperation. As expected, BCO performs better in this case than from translated demonstrations, however the performance is still significantly worse than \metabbr\ from translated demonstrations. This indicates that learning from full demonstrations, even directly obtained from the robot, suffers from accumulating errors for long-horizon tasks. We note that, to our knowledge, BCO has never been tested on multi-stage tasks from image observations~\cite{bco}.

Finally, behavioral cloning outperforms our method for coffee making but surprisingly performs worse on cup retrieval. We believe that this is because the cup retrieval task has more stages and is harder to learn without explicit stage-wise training. Thus, compared to \metabbr, behavioral cloning has more stringent assumptions and may perform worse for more complex tasks, but the drawback of \metabbr\ is that it uses an RL training phase that requires human feedback. To understand the human burden associated with this training procedure, we plot the human feedback we provide during training in \autoref{fig:humanfeedback}. Training took about one hour for each task, and in both cases \metabbr\ used less than 150 key presses to learn the task, which was easy for a human supervisor to provide. We view this as a practical way to enable robots to learn complex tasks.

\section{Discussion and Future Work}

We presented \metabbr, a method for learning visual robotic multi-stage tasks directly from videos of human demonstrations. \metabbr\ uses image-to-image translation, via CycleGAN, to generate robot instruction images that enable a stage-wise model-based RL algorithm to learn from raw image observations. The RL algorithm is based on latent space model-based planning and incorporates learned resets and human feedback to learn temporally extended tasks robustly and efficiently. We demonstrated that \metabbr\ is capable of learning multi-stage coffee making and cup retrieval tasks, attains substantially better results than prior methods and ablations using human demonstrations, and even outperforms behavioral cloning using teleoperated demonstrations (rather than videos of human demonstrations) on one of the tasks.

One advantage of our framework is that, in principle, the translation module is decoupled from the RL process and is not limited to a particular task. Thus, the most exciting direction for future work is to amortize the cost of data collection and CycleGAN training across multiple tasks, rather than training separate CycleGANs for each task of interest. For example, the cup retrieval CycleGAN can also be used to translate demonstrations for placing the cup back in the drawer, meaning that this new task can be learned from a handful of human demonstrations with no additional upfront cost. As an initial step toward this goal, we experimented with training a single CycleGAN on data collected from both the coffee making and cup retrieval tasks, the results of which can be found on the project website.\footnote{\tt \url{https://sites.google.com/view/rss20avid}} We aim to further generalize this result by training a CycleGAN on an initial large dataset, e.g., many different human and robot behaviors in a kitchen that has a coffee machine, multiple drawers, and numerous other objects. This should enable any new task in the kitchen to be learned with just a few human demonstrations of the task, and this is a promising direction toward truly allowing robots to learn by watching humans.
\subsection*{Acknowledgements}
This work was supported in part by Berkeley Deep Drive, NSF IIS-1700696 and IIS-1651843, and ONR PECASE N000141612723 and ONR N000141912042. MZ was supported by an NDSEG Fellowship.

\balance

\bibliographystyle{IEEEtran}
\bibliography{references}

\begin{thebibliography}{10}
\providecommand{\url}[1]{#1}
\csname url@samestyle\endcsname
\providecommand{\newblock}{\relax}
\providecommand{\bibinfo}[2]{#2}
\providecommand{\BIBentrySTDinterwordspacing}{\spaceskip=0pt\relax}
\providecommand{\BIBentryALTinterwordstretchfactor}{4}
\providecommand{\BIBentryALTinterwordspacing}{\spaceskip=\fontdimen2\font plus
\BIBentryALTinterwordstretchfactor\fontdimen3\font minus
  \fontdimen4\font\relax}
\providecommand{\BIBforeignlanguage}[2]{{%
\expandafter\ifx\csname l@#1\endcsname\relax
\typeout{** WARNING: IEEEtran.bst: No hyphenation pattern has been}%
\typeout{** loaded for the language `#1'. Using the pattern for}%
\typeout{** the default language instead.}%
\else
\language=\csname l@#1\endcsname
\fi
#2}}
\providecommand{\BIBdecl}{\relax}
\BIBdecl

\bibitem{robot-lfd}
B.~Argall, S.~Chernova, M.~Veloso, and B.~Browning, ``A survey of robot
  learning from demonstration,'' \emph{Robotics and Autonomous Systems}, 2009.

\bibitem{collab-manip}
S.~Calinon, P.~Evrard, E.~Gribovskaya, A.~Billard, and A.~Kheddar, ``Learning
  collaborative manipulation tasks by demonstration using a haptic interface,''
  in \emph{ICAR}, 2009.

\bibitem{online-movement-adaptation}
P.~Pastor, L.~Righetti, M.~Kalakrishnan, and S.~Schaal, ``Online movement
  adaptation based on previous sensor experiences,'' in \emph{IROS}, 2011.

\bibitem{traj-keyframe}
B.~Akgun, M.~Cakmak, J.~Yoo, and A.~Thomaz, ``Trajectories and keyframes for
  kinesthetic teaching: A human-robot interaction perspective,'' in \emph{HRI},
  2012.

\bibitem{vr-teleop}
T.~Zhang, Z.~McCarthy, O.~Jow, D.~Lee, K.~Goldberg, and P.~Abbeel, ``Deep
  imitation learning for complex manipulation tasks from virtual reality
  teleoperation,'' in \emph{ICRA}, 2018.

\bibitem{cyclegan}
J.~Zhu, T.~Park, P.~Isola, and A.~Efros, ``Unpaired image-to-image translation
  using cycle-consistent adversarial networks,'' in \emph{ICCV}, 2017.

\bibitem{quadrupedal}
N.~Kohl and P.~Stone, ``Policy gradient reinforcement learning for fast
  quadrupedal locomotion,'' in \emph{ICRA}, 2004.

\bibitem{policy-search}
J.~Kober and J.~Peters, ``Policy search for motor primitives in robotics,''
  \emph{Machine Learning}, 2011.

\bibitem{robot-rl}
J.~Kober, J.~Bagnell, and J.~Peters, ``Reinforcement learning in robotics: A
  survey,'' \emph{IJRR}, 2013.

\bibitem{tactile-features}
H.~van Hoof, T.~Hermans, G.~Neumann, and J.~Peters, ``Learning robot in-hand
  manipulation with tactile features,'' in \emph{Humanoids}, 2015.

\bibitem{in-hand-manip}
{OpenAI} \emph{et~al.}, ``Learning dexterous in-hand manipulation,''
  \emph{arXiv preprint arXiv:1808.00177}, 2018.

\bibitem{qtopt}
D.~Kalashnikov, A.~Irpan, P.~Pastor, J.~Ibarz, A.~Herzog, E.~Jang, D.~Quillen,
  E.~Holly, M.~Kalakrishnan, V.~Vanhoucke, and S.~Levine, ``{QT-O}pt: Scalable
  deep reinforcement learning for vision-based robotic manipulation,'' in
  \emph{CoRL}, 2018.

\bibitem{em-rl}
P.~Kormushev, S.~Calinon, and D.~Caldwell, ``Robot motor skill coordination
  with {EM}-based reinforcement learning,'' in \emph{IROS}, 2010.

\bibitem{progressive-nets}
A.~Rusu, M.~Ve\u{c}er\'ik, T.~Roth\"orl, N.~Heess, R.~Pascanu, and R.~Hadsell,
  ``Sim-to-real robot learning from pixels with progressive nets,'' in
  \emph{CoRL}, 2017.

\bibitem{dagps}
A.~Yahya, A.~Li, M.~Kalakrishnan, Y.~Chebotar, and S.~Levine, ``Collective
  robot reinforcement learning with distributed asynchronous guided policy
  search,'' in \emph{IROS}, 2017.

\bibitem{pouring}
C.~Schenck and D.~Fox, ``Visual closed-loop control for pouring liquids,'' in
  \emph{ICRA}, 2017.

\bibitem{srl-survey}
T.~Lesort, N.~D\'{i}az-Rodr\'{i}guez, J.~Goudou, and D.~Filliat, ``State
  representation learning for control: An overview,'' \emph{Neural Networks},
  2018.

\bibitem{pilco}
M.~Deisenroth, D.~Fox, and C.~Rasmussen, ``Gaussian processes for
  data-efficient learning in robotics and control,'' \emph{PAMI}, 2014.

\bibitem{pets}
K.~Chua, R.~Calandra, R.~McAllister, and S.~Levine, ``Deep reinforcement
  learning in a handful of trials using probabilistic dynamics models,'' in
  \emph{NIPS}, 2018.

\bibitem{slac}
A.~Lee, A.~Nagabandi, P.~Abbeel, and S.~Levine, ``Stochastic latent
  actor-critic: Deep reinforcement learning with a latent variable model,''
  \emph{arXiv preprint arXiv:1907.00953}, 2019.

\bibitem{vice}
J.~Fu, A.~Singh, D.~Ghosh, L.~Yang, and S.~Levine, ``Variational inverse
  control with events: A general framework for data-driven reward definition,''
  in \emph{NIPS}, 2018.

\bibitem{seq-mp}
F.~Stulp, E.~Theodorou, and S.~Schaal, ``Reinforcement learning with sequences
  of motion primitives for robust manipulation,'' \emph{IEEE Transactions on
  Robotics}, 2012.

\bibitem{reset-controller}
W.~Han, S.~Levine, and P.~Abbeel, ``Learning compound multi-step controllers
  under unknown dynamics,'' in \emph{IROS}, 2015.

\bibitem{rope-manip}
A.~Nair, D.~Chen, P.~Agrawal, P.~Isola, P.~Abbeel, J.~Malik, and S.~Levine,
  ``Combining self-supervised learning and imitation for vision-based rope
  manipulation,'' in \emph{ICRA}, 2017.

\bibitem{htpi}
P.~Sharma, D.~Pathak, and A.~Gupta, ``Third-person visual imitation learning
  via decoupled hierarchical control,'' in \emph{NeurIPS}, 2019.

\bibitem{mp-demo}
S.~Manschitz, J.~Kober, M.~Gienger, and J.~Peters, ``Leraning to sequence
  movement primitives from demonstrations,'' in \emph{IROS}, 2014.

\bibitem{causal-infogan}
T.~Kurutach, A.~Tamar, G.~Yang, S.~Russell, and P.~Abbeel, ``Learning plannable
  representations with causal info{GAN},'' \emph{arXiv preprint
  arXiv:1807.09341}, 2018.

\bibitem{active}
C.~Daniel, M.~Viering, J.~Metz, O.~Kroemer, and J.~Peters, ``Active reward
  learning,'' in \emph{RSS}, 2014.

\bibitem{human-pref}
P.~Christiano, J.~Leike, T.~Brown, M.~Martic, S.~Legg, and D.~Amodei, ``Deep
  reinforcement learning from human preferences,'' in \emph{NIPS}, 2017.

\bibitem{vice-raq}
A.~Singh, L.~Yang, K.~Hartikainen, C.~Finn, and S.~Levine, ``End-to-end robotic
  reinforcement learning without reward engineering,'' in \emph{RSS}, 2019.

\bibitem{learning-imitation}
G.~Hayes and J.~Demiris, ``A robot controller using learning by imitation,''
  1994.

\bibitem{human-arm}
A.~Billard and M.~Mataric, ``Learning human arm movements by imitation:
  Evaluation of a biologically inspired connectionist architecture,''
  \emph{Robotics and Autonomous Systems}, 2001.

\bibitem{comp-approaches}
S.~Schaal, A.~Ijspeert, and A.~Billard, ``Computational approaches to motor
  learning by imitation,'' \emph{Philosophical Transaction of the Royal Society
  of London, Series B}, 2003.

\bibitem{mp-robot}
J.~Kober and J.~Peters, ``Learning motor primitives for robotics,'' in
  \emph{ICRA}, 2009.

\bibitem{obj-functions}
M.~Kalakrishnan, P.~Pastor, L.~Righetti, and S.~Schaal, ``Learning objective
  functions for manipulation,'' in \emph{ICRA}, 2013.

\bibitem{kendama}
H.~Miyamoto, S.~Schaal, F.~Gandolfo, H.~Gomi, Y.~Koike, R.~Osu, E.~Nakano,
  Y.~Wada, and M.~Kawato, ``A {K}endama learning robot based on bi-directional
  theory,'' \emph{Neural Networks}, 1996.

\bibitem{learning-watching}
Y.~Kuniyoshi, M.~Inaba, and H.~Inoue, ``Learning by watching: Extracting
  reusable task knowledge from visual observation of human performance,''
  \emph{IEEE Transactions on Robotics and Automation}, 1994.

\bibitem{activity-grammars}
K.~Lee, Y.~Su, T.~Kim, and Y.~Demiris, ``A syntactic approach to robot
  imitation learning using probabilistic activity grammars,'' \emph{Robotics
  and Autonomous Systems}, 2013.

\bibitem{world-wide-web}
Y.~Yang, Y.~Li, C.~Fermuller, and Y.~Aloimonos, ``Robot learning manipulation
  action plans by ``watching'' unconstrained videos from the world wide web,''
  in \emph{AAAI}, 2015.

\bibitem{semantic-rep}
K.~Ramirez-Amaro, M.~Beetz, and G.~Cheng, ``Transferring skills to humanoid
  robots by extracting semantic representations from observations of human
  activities,'' \emph{Artificial Intelligence}, 2017.

\bibitem{correspondence}
C.~Nehaniv and K.~Dautenhahn, ``The correspondence problem,'' \emph{Imitation
  in Animals and Artifacts}, 2002.

\bibitem{activity-forecasting}
N.~Rhinehart and K.~Kitani, ``First-person activity forecasting with online
  inverse reinforcement learning,'' in \emph{ICCV}, 2017.

\bibitem{wwyd}
A.~Tow, N.~S\"{u}nderhauf, S.~Shirazi, M.~Milford, and J.~Leitner, ``What would
  you do? {A}cting by learning to predict,'' in \emph{IROS}, 2017.

\bibitem{ifo}
Y.~Liu, A.~Gupta, P.~Abbeel, and S.~Levine, ``Imitation from observation:
  Learning to imitate behaviors from raw video via context translation,'' in
  \emph{ICRA}, 2018.

\bibitem{perceptual-rewards}
P.~Sermanet, K.~Xu, and S.~Levine, ``Unsupervised perceptual rewards for
  imitation learning,'' in \emph{RSS}, 2017.

\bibitem{tcn}
P.~Sermanet, C.~Lynch, Y.~Chebotar, J.~Hsu, E.~Jang, S.~Schaal, and S.~Levine,
  ``Time-contrastive networks: Self-supervised learning from video,'' in
  \emph{ICRA}, 2018.

\bibitem{daml}
T.~Yu, C.~Finn, A.~Xie, S.~Dasari, P.~Abbeel, and S.~Levine, ``One-shot
  imitation from observing humans via domain-adaptive meta-learning,'' in
  \emph{RSS}, 2018.

\bibitem{unit}
M.~Liu, T.~Breuel, and J.~Kautz, ``Unsupervised image-to-image translation
  networks,'' in \emph{NIPS}, 2017.

\bibitem{solar}
M.~Zhang, S.~Vikram, L.~Smith, P.~Abbeel, M.~Johnson, and S.~Levine, ``{SOLAR}:
  Deep structured representations for model-based reinforcement learning,'' in
  \emph{ICML}, 2019.

\bibitem{planet}
D.~Hafner, T.~Lillicrap, I.~Fischer, R.~Villegas, D.~Ha, H.~Lee, and
  J.~Davidson, ``Learning latent dynamics for planning from pixels,'' in
  \emph{ICML}, 2018.

\bibitem{zsvi}
D.~Pathak, P.~Mahmoudieh, G.~Luo, P.~Agrawal, D.~Chen, Y.~Shentu, E.~Shelhamer,
  J.~Malik, A.~Efros, and T.~Darrell, ``Zero-shot visual imitation,'' in
  \emph{ICLR}, 2018.

\bibitem{selective-il}
Y.~Lee, E.~Hu, Z.~Yang, and J.~Lim, ``To follow or not to follow: Selective
  imitation learning from observations,'' \emph{arXiv preprint
  arXiv:1912.07670}, 2019.

\bibitem{tap}
D.~Jayaraman, F.~Ebert, A.~Efros, and S.~Levine, ``Time-agnostic prediction:
  Predicting predictable video frames,'' in \emph{ICLR}, 2019.

\bibitem{tsc}
S.~Krishnan, A.~Garg, S.~Patil, C.~Lea, G.~Hager, P.~Abbeel, and K.~Goldberg,
  ``Transition state clustering: Unsupervised surgical trajectory segmentation
  for robot learning,'' \emph{IJRR}, 2017.

\bibitem{cem}
R.~Rubinstein and D.~Kroese, \emph{The Cross Entropy Method: A Unified Approach
  To Combinatorial Optimization, {M}onte-{C}arlo Simulation (Information
  Science and Statistics)}.\hskip 1em plus 0.5em minus 0.4em\relax
  Springer-Verlag New York, Inc., 2004.

\bibitem{bco}
F.~Torabi, G.~Warnell, and P.~Stone, ``Behavioral cloning from observation,''
  in \emph{IJCAI}, 2018.

\bibitem{vf}
F.~Ebert, C.~Finn, S.~Dasari, A.~Xie, A.~Lee, and S.~Levine, ``Visual
  foresight: Model-based deep reinforcement learning for vision-based robotic
  control,'' \emph{arXiv preprint arXiv:1812.00568}, 2018.

\bibitem{dagger}
S.~Ross, G.~Gordon, and J.~Bagnell, ``A reduction of imitation learning and
  structured prediction to no-regret online learning,'' in \emph{AISTATS},
  2011.

\bibitem{fail}
W.~Sun, A.~Vemula, B.~Boots, and J.~Bagnell, ``Provably efficient imitation
  learning from observation alone,'' in \emph{ICML}, 2019.

\bibitem{ilpo}
A.~Edwards, H.~Sahni, Y.~Schroecker, and C.~Isbell, ``Imitating latent policies
  from observation,'' in \emph{ICML}, 2019.

\bibitem{pilqr}
Y.~Chebotar, K.~Hausman, M.~Zhang, G.~Sukhatme, S.~Schaal, and S.~Levine,
  ``Combining model-based and model-free updates for trajectory-centric
  reinforcement learning,'' in \emph{ICML}, 2017.

\end{thebibliography}

\end{document}